\documentclass{article}

   \usepackage[preprint]{tackling_climate_workshop_style}

\usepackage[utf8]{inputenc} % allow utf-8 input
\usepackage[T1]{fontenc}    % use 8-bit T1 fonts
\usepackage{hyperref}       % hyperlinks
\usepackage{url}            % simple URL typesetting
\usepackage{booktabs}       % professional-quality tables
\usepackage{amsfonts}       % blackboard math symbols
\usepackage{nicefrac}       % compact symbols for 1/2, etc.
\usepackage{microtype}      % microtypography
\usepackage{graphicx}

\title{Precision Agriculture: Crop Mapping using Machine Learning and Sentinel-2 Satellite Imagery}

\author{%
  Kui Zhao \\
  Queen Mary University of London\\
  London, UK\\
  \texttt{k.zhao.uk@outlook.com} \\
  \And
  Siyang Wu  \\
  Queen Mary University of London \\
  London, UK \\
  \texttt{wusy950320@163.com} \\
  \And
  Chang Liu \\
  Queen Mary University of London \\
   London, UK \\
   \texttt{carolliu21573@163.com} \\
  \And
  Yue Wu \\
  Queen Mary University of London \\
   London, UK \\
   \texttt{1476143875@qq.com} \\
  \And 
  Natalia Efremova\\
  Queen Mary University of London\\
  London, UK \\
  \texttt{n.efremova@qmul.ac.uk} \\
}

\begin{document}

\maketitle

\begin{abstract}

Food security has grown in significance due to the changing climate and its warming effects. To support the rising demand for agricultural products and to minimize the negative impact of climate change and mass cultivation, precision agriculture has become increasingly important for crop cultivation. This study employs deep learning and pixel-based machine learning methods to accurately segment lavender fields for precision agriculture, utilizing various spectral band combinations extracted from Sentinel-2 satellite imagery. Our fine-tuned final model, a U-Net architecture, can achieve a Dice coefficient of 0.8324. Additionally, our investigation highlights the unexpected efficacy of the pixel-based method and the RGB spectral band combination in this task.
\end{abstract}

\section{Introduction}
\label{sec1:intro}
The primary reason for our interest in applying artificial intelligence (AI) to agriculture or studying agriculture within the field of AI, and the reason why we discuss agriculture in relation to climate change AI, is the critical issue of food security. Food security is widely recognized as one of the most urgent challenges we currently face globally [1]. This concern has grown in significance due to the changing climate and its warming effects. With the increasing occurrence of extreme weather events and alterations in weather patterns, our food system, especially in certain regions, has become highly vulnerable to the impacts of climate change. Our key objective is to develop more reliable and scalable methods for monitoring global crop conditions promptly and transparently, while also exploring how we can adapt agriculture to mitigate the effects of climate change. To address this objective, we focus on the case of lavender cultivation. Its cultivation is affected by climate change and it holds substantial economic value in industries such as aromatherapy, cosmetics, and food [2,3]. Its demand has been steadily rising alongside population growth [4], taking up significant resources, and causing environmental impact [5,6]. 

In this study, we collected remote sensing data from the Sentinel-2 satellite and used machine learning to segment the lavender field for timely monitoring of its condition. This is pivotal in adapting its traditional agricultural practices to precision agriculture to mitigate the effects of climate change [7]. This transition maintains or improves production while reducing resource usage and negative environmental impact.

Previous works used different band combinations ranging from RGB to multispectral bands [8,9] to employ machine learning models including U-Net and SegNet [10-16]. In this study, we investigate the performance of different model-spectral band combinations to comprehensively assess the application of machine learning in monitoring lavender conditions using remote sensing imagery.
 
\section{Methods}
\label{sec2:methods}
\subsection{Data processing}
We collected 13 Sentinel-2 L2A satellite imagery on 12 spectral bands, with varying resolution and size across bands and images. Among these images, approximately 16.22\% of the pixels are annotated as lavender fields, leaving the other 83.78\% to represent the background. This distribution is slightly imbalanced. We plotted the images in their RGB bands and found that the lavender fields during the blooming period give out a distinctive purple hue, which makes them easily detectable. We resized the images to a size that can be patched into 96x96-pixel-size patches and patched them. We then excluded empty patches to mitigate the imbalance issue and ensure the accuracy of our evaluation metrics, such as the Dice coefficient and intersection over union (IoU). 

Normalized Difference Vegetation Index (NDVI) is a spectral imaging transformation of near-infrared and red bands designed to quantify the amount of vegetation. Normalized Difference Moisture Index (NDMI) is a transformation of short-wave infrared and near-infrared bands designed to detect moisture levels in vegetation. They are commonly employed for image segmentation of crops to enhance the model's performance, so they are stacked with the original 12 bands of data.  

\subsection{Model architecture}
\label{subsec:model_arch}

We used 2 methods for image segmentation. One is the deep learning method, which includes models like U-Net [12], SegNet [13], Unet++ [14] and ResU-Net [15]. The other method is the pixel-based machine learning method. They take each pixel’s value in different bands as the input features and use classic machine learning models such as Logistic Regression, Decision Tree, and Random Forest [16] to do binary classification. They don’t take into account the relationships between pixels. After comparison, we chose U-Net for hyperparameter tuning. We also tested different methods’ performance under different combinations of spectral bands.

\section{Results and discussion}
\label{sec:results}

\begin{table}
  \caption{Performance of different models and model selection. Deep learning method (DL), pixel-based machine learning method (ML), with Gradient Clipping (w/GC).}
  \label{table:performance}
  \centering
  \begin{tabular}{lllll}
    \toprule
    Method     & Model     & F1 score & Dice coefficient & IoU\\
    \midrule
    DL & SegNet [13]     & 0.8460  & 0.8460  & 0.7331  \\
    \textbf{DL} & \textbf{U-Net [12]}      & \textbf{0.8615} & \textbf{0.8615} & \textbf{0.7568}  \\
    DL & U-Net w/GC & 0.8425 & 0.8425 & 0.7279  \\
    DL & U-Net++ [14]    & 0.8537 & 0.8537 & 0.7447  \\
    DL & ResU-Net [15]   & 0.7973 & 0.7973 & 0.6629  \\
    ML & Logistic Reg     & 0.8344  & 0.8344  &0.7159  \\
    ML & Decision Tree & 0.8146 & 0.8146 & 0.6872  \\
    ML & Random Forest [16]  & 0.8720 & 0.8720 & 0.7731  \\
    \bottomrule
  \end{tabular}
\end{table}

\subsection{Deep Learning Method}
\label{subsec:dl}
As is shown by Table 1, U-Net, U-Net++, and SegNet can reach a very high Dice coefficient of 0.8615, 0.8537, and 0.8460 respectively, which is almost the same with slight differences due to the random nature of neural network training. Although these scores are slightly below the Random Forest's Dice coefficient of 0.8720, they still represent a robust level of segmentation accuracy. This suggests that for this task the basic U-Net is able to capture most of the patterns that more sophisticated models can capture. Therefore, we chose U-Net for hyperparameter tuning for its simplicity and generalizability.

From Figure 1, we see that our DL models tend to overfit. Also, their loss suddenly increases during training, resulting in worse and unstable performance which can cause such problems as unpredictable training behavior. We would solve these issues by adding regularization term and dropout layer.

\begin{figure}
  \centering
  \includegraphics[width = 8cm]{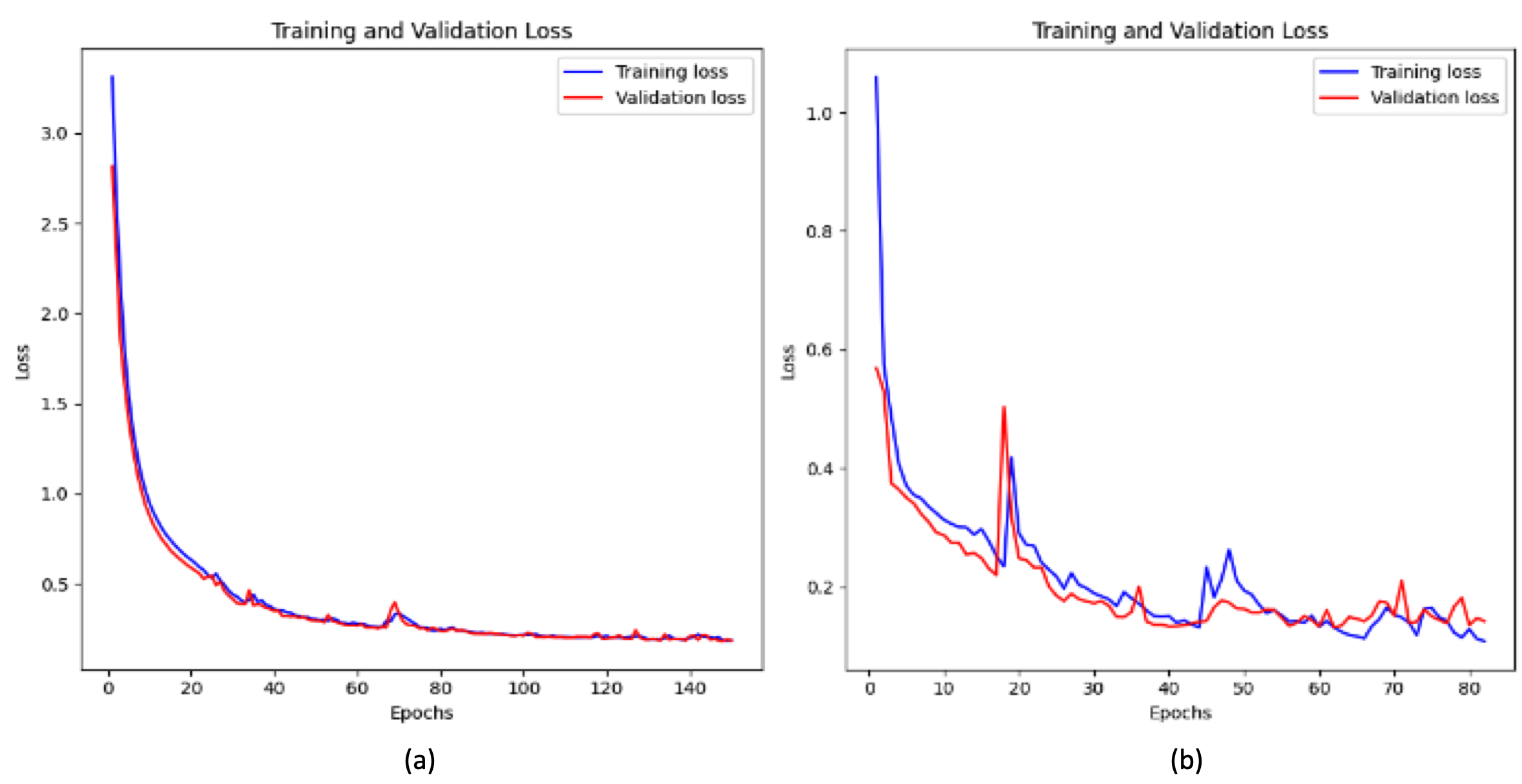}
  \caption{Learning curve of tuned U-Net model(a) and default U-Net model(b).}
  \label{fig:train_curve}
\end{figure}

\subsection{Pixel-based Method}
\label{subsec:pix}
We find that classic ML methods, such as logistic regression, demonstrate great performance on the combination of 12 bands plus two added indexes,  even surpassing the performance of DL models. Notably, the Logistic Regression demonstrates a remarkable performance comparable with its deep learning counterparts such as U-Net. This is particularly noteworthy because Logistic Regression’s training time is mere seconds compared to the incredibly long training time of deep learning models. This highlights the potential for well-selected classic models for segmentation tasks, especially in cases where data availability allows for effective learning.

\begin{figure}
  \centering
  \includegraphics[width = 10cm]{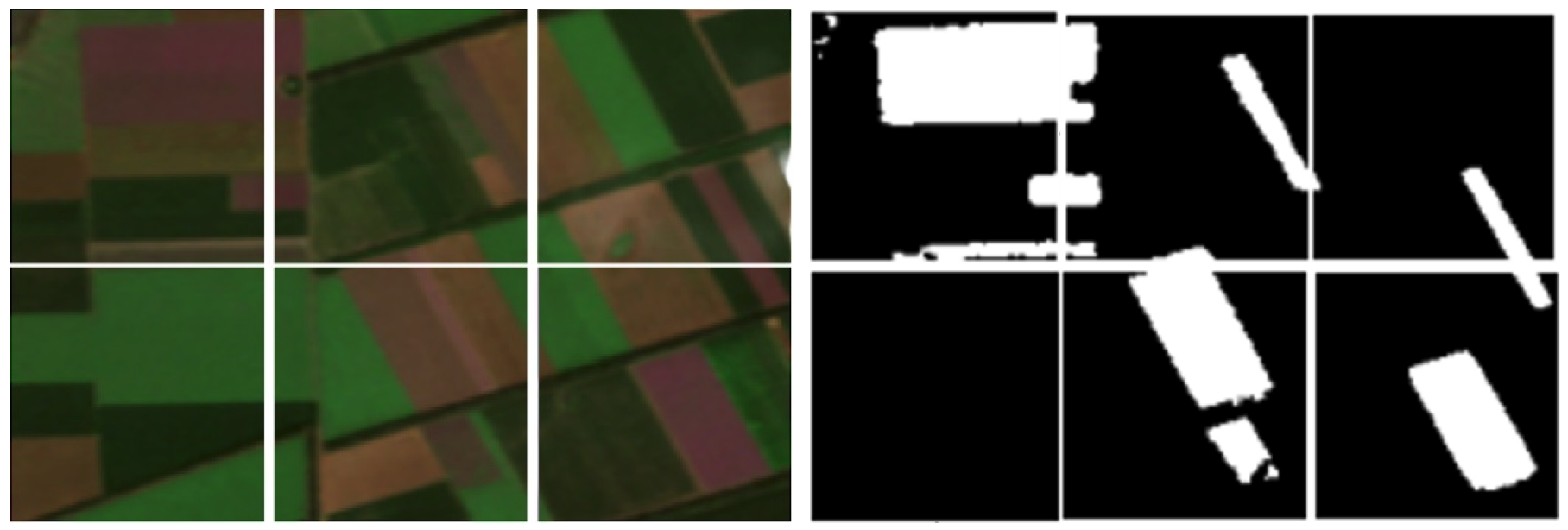}
  \caption{Patched sample image in RGB and the predicted output for individual patches.}
  \label{fig:patches}
\end{figure}

\begin{figure}
  \centering
  \includegraphics[width = 10cm]{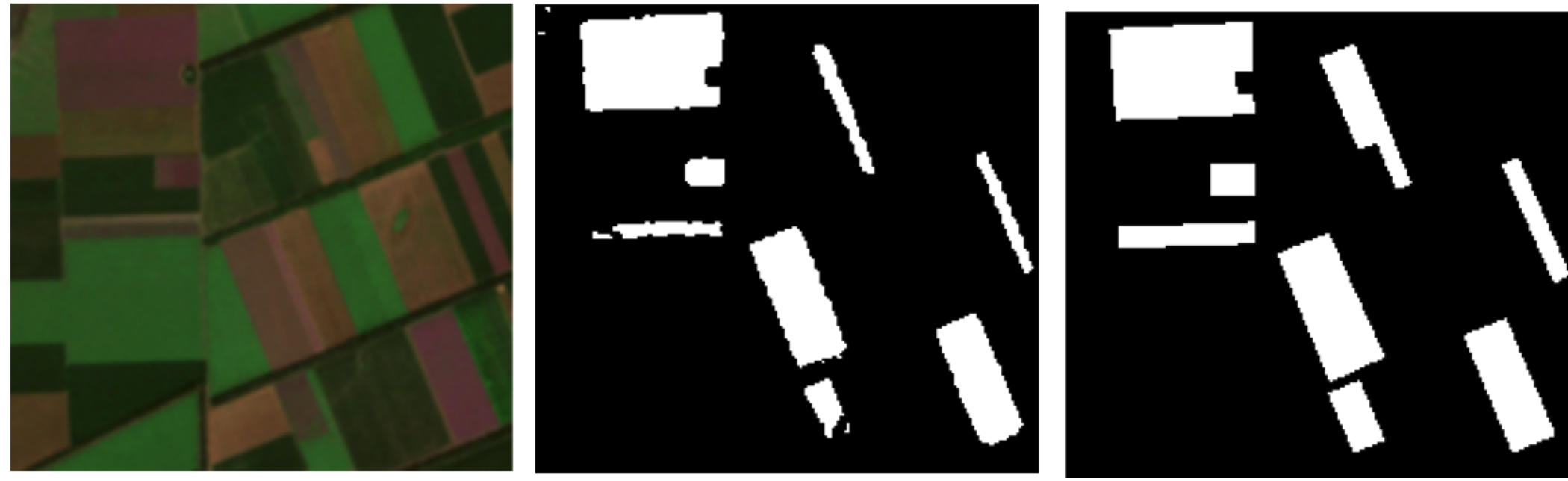}
  \caption{RGB image(a), predicted output(b), and ground truth mask (c).}
  \label{fig:mask_comparison}
\end{figure}

\subsection{Hyper-parameter tuning and the predicted mask}
\label{subsec:hyper}

After tuning, the final model is a U-Net architecture consisting of four downsampling-upsampling stages. It takes as an input 96x96-pixel images with 14 channels (12 spectral bands and 2 indexes) to produce binary masks. The model's encoder utilizes convolutional layers with max-pooling and dropout for dimension reduction, followed by a bottleneck layer with additional convolutions. The decoder incorporates up-sampling, convolutional layers, and concatenation of encoder features. The output layer generates binary segmentation masks with sigmoid activation. Training employs binary cross-entropy loss and early stopping to prevent overfitting. Key hyperparameters include a dropout rate of 0.1, a learning rate of 0.001, L2 regularization with a strength of 0.001, and weight initialization using the 'lecun\_normal' strategy. 

Table 2 and Figure 1 demonstrate the comparison of the U-net model performance before and after tuning, including improvement in performance and stability during training. The fine-tuned U-Net model can achieve a Dice coefficient of 0.8324 on the test set. After training the model, the predicted masks for individual patches are combined and resized to their original image as shown in Figure 2 and 3. The predicted mask and the ground truth for the original satellite image are almost the same, suggesting the success of our segmentation task.

\begin{table}
  \caption{The average performance of the default and tuned U-Net Model over 10 times training. All models are trained using the following hyper-parameters: number epochs: 150; batch size: 16; patience: 20; Loss function: cross-entropy.}
  \label{table:unet}
  \centering
  \begin{tabular}{lllllll}
    \toprule
    Model     & Dropout rate & Weight initialisation & L2 & Avg.F1& Avg.Dice  & Avg.IoU\\
    \midrule
    U-Net tuned & 0.1 & lecun\_normal   &0.001 & 0.8324	& 0.8324	&0.7129\\
    U-Net       & 0 & glorot\_uniform &None  & 0.8028    & 0.8028	&0.6709\\   
    \bottomrule
  \end{tabular}
\end{table}

\subsection{Model performance under different spectral band combinations}
\label{subsec:perf}

\begin{table}
  \caption{Model performance (Dice coefficient) under different spectral band combinations. RGB: the red, green, and blue band combination; SWIR: the short-wave infrared band combination; CIR: the color infrared combination; All 12: first 12 spectral band combination from Sentinel-2 L2A}
  \label{table:bands}
  \centering
  \begin{tabular}{llllllllll}
    \toprule
    Method & Model & RGB  & SWIR & CIR & All 12 & All 12 + NDVI + NDMI \\
    \midrule
    DL & U-Net tuned & 0.8458  & 0.7752 & 0.7363 & 0.8685 & 0.8350 \\
    DL & U-Net & 0.8314  & 0.7801 & 0.6785 & 0.8434 & 0.8146 \\
    ML & Logistic R & 0.7249  & 0.3601 & 0.3788 & 0.8347 & 0.8534 \\
    ML & Decision Tree & 0.7597  & 0.6905 & 0.6782 & 0.8166 & 0.8334 \\
    \bottomrule
  \end{tabular}
\end{table}

As is shown by Table 3, the U-Net models' performance using only the RGB bands is almost on par with utilizing the full set of 12 bands. This suggests that the essential information required for accurate segmentation can be captured through RGB alone. U-Net models achieve higher performance than pixel-based ML methods, especially in SWIR and CIR combinations. 
The performance of the pixel-based ML method experiences a sharp decline when employing the RGB alone instead of the complete 12-band dataset. Additional bands demonstrate improved performance, highlighting the ML method's reliance on a more comprehensive set of bands. 

Based on our results, we propose that in scenarios where 12 or more bands are available for image segmentation and a slight performance decrease is acceptable, pixel-based ML methods such as logistic regression can be a cost-effective and time-efficient choice. But U-Net performs better in almost every scenario, though computationally expensive. Using U-net, just the RGB image would give enough information for effective segmentation of the lavender field.

\section{Conclusions}
In conclusion, our tuned U-Net model can reach a 0.8324 Dice coefficient for the lavenders. The pixel-based machine learning can also reach a high Dice coefficient, but it is reliant on multispectral data from more sensors. In comparison, U-Net can reach a high performance with RGB spectral bands alone. Our findings provide valuable support for the lavender industry's adoption of precision agriculture practices in response to climate change.

\section*{References}
\medskip
\small
[1] Lee, H., \ \&  Calvin, K., \ \&  Dasgupta, D., \ \&  Krinner, G., \ \&  Mukherji, A., \ \&  Thorne, P., et al. \ (2023). { \it AR6 Synthesis Report: Climate Change 2023}. Summary for Policymakers.

[2] Lee, I. S., \ \& Lee, G. J. \ (2006). {\it Effects of lavender aromatherapy on insomnia and depression in women college students}. Journal of Korean Academy of Nursing, 36(1), 136-143.

[3] Wells, R., \ \&  Truong, F., \ \&  Adal, A. M.,\ \&  Sarker, L. S., \ \&  Mahmoud, S. S. \ (2018). {\it Lavandula essential oils: a current review of applications in medicinal, food, and cosmetic industries of lavender}. Natural Product Communications, 13(10).

[4] Crișan, I.,\ \& Ona, A.,\ \& Vârban, D.,\ \& Muntean, L., et al. \ (2023). {\it Current trends for lavender (lavandula angustifolia Mill.) crops and products with emphasis on essential oil quality}. Plants, 12(2), 357.

[5] Rodríguez Pleguezuelo, C. R., \ \& Durán Zuazo, V. H., \ \& Martínez Raya, A., \ \& Francia Martínez, J. R., \ \& Cárceles Rodríguez, B. \ (2009). {\it High reduction of erosion and nutrient losses by decreasing harvest intensity of lavender grown on slopes}. Agronomy for sustainable development, 29(2), 363-370.

[6] Lilienfeld, A., \ \&  Asmild, M. \ (2007). {\it Estimation of excess water use in irrigated agriculture: A Data Envelopment Analysis approach}. Agricultural water management, 94(1-3), 73-82.

[7] Pierce, F. J., \ \&  Nowak, P. \ (1999). {\it Aspects of precision agriculture}. Advances in agronomy, 67, 1-85.

[8] Liebisch, F.,\ \&  Kirchgessner, N., \ \& Schneider, D., Walter, A., \ \&  Hund, A. \ (2015). {\it Remote, aerial phenotyping of maize traits with a mobile multi-sensor approach}. Plant methods, 11(1), 1-20.

[9] Sa, I.,\ \&  Popović, M.,\ \&  Khanna, R.,\ \&  Chen, Z.\ \& , Lottes, P., et al. \ (2018). {\it WeedMap: A large-scale semantic weed mapping framework using aerial multispectral imaging and deep neural network for precision farming}. Remote Sensing, 10(9), 1423.

[10] Fawakherji, M.,\ \&  Youssef, A.,\ \&  Bloisi, D.,\ \&  Pretto, A., \ \&  Nardi, D. \ (2019). {\it Crop and weeds classification for precision agriculture using context-independent pixel-wise segmentation}. In 2019 Third IEEE International Conference on Robotic Computing (IRC) (pp. 146-152). IEEE.

[11] Kerkech, M.,\ \&  Hafiane, A., \ \&  Canals, R. \ (2020). {\it Vine disease detection in UAV multispectral images using optimized image registration and deep learning segmentation approach}. Computers and Electronics in Agriculture, 174, 105446.

[12] Ronneberger, O.,\ \&  Fischer, P., \ \&  Brox, T. \ (2015). {\it U-net: Convolutional networks for biomedical image segmentation}. MICCAI 2015: 18th International Conference, Munich, Germany, October 5-9, 2015, Proceedings, Part III 18 (pp. 234-241). Springer International Publishing.

[13] Badrinarayanan, V.,\ \&  Kendall, A., \ \&  Cipolla, R. (2017). {\it Segnet: A deep convolutional encoder-decoder architecture for image segmentation}. IEEE transactions on pattern analysis and machine intelligence, 39(12), 2481-2495.

[14] Zhou, Z.,\ \&  Rahman Siddiquee, M. M.,\ \&  Tajbakhsh, N., \ \&  Liang, J. \ (2018). { \it Unet++: A nested u-net architecture for medical image segmentation}. MICCAI 2018, Granada, Spain, September 20, 2018, Proceedings 4 (pp. 3-11). Springer International Publishing.

[15] Diakogiannis, F. I.,\ \&  Waldner, F.,\ \&  Caccetta, P.,\ \&  Wu, C. \ (2020). { \it ResUNet-a: A deep learning framework for semantic segmentation of remotely sensed data}. ISPRS Journal of Photogrammetry and Remote Sensing, 162, 94-114.

[16] Breiman, L. \ (2001) { \it Random forests}. In Machine Learning, 5–32

\end{document}